\documentclass[lettersize,journal]{IEEEtran}
\usepackage{amsmath,amsfonts}
\usepackage[boxed,ruled]{algorithm2e}
\usepackage{array}
\usepackage[caption=false,font=normalsize,labelfont=sf,textfont=sf]{subfig}
\usepackage{textcomp}
\usepackage{stfloats}
\usepackage{url}
\usepackage{verbatim}
\usepackage{graphicx}
\usepackage{amssymb}
\usepackage{cite}
\usepackage{booktabs}
\usepackage{balance}
\usepackage{pifont}
\usepackage{utfsym}
\usepackage{multirow} 
\usepackage[dvipsnames]{xcolor} 

\usepackage{tabularx, makecell, multirow} 
\newcommand{\cz}[0]{\textcolor{black}}

\begin{document}

\title{Synergistic Modality-and-Slice Memory Framework for Cross-Modal 3D Brain Tumor Segmentation}

\author{Yuxiang Luo, Qing Xu, Hai Huang, Yuqi Ouyang, Xiangjian He, \IEEEmembership{Senior Member, IEEE}, \\Zhen Chen, Wenting Duan, Jiebo Luo, \IEEEmembership{Fellow, IEEE}  
        \thanks{\quad {This work is partially supported by the Program of China Scholarship Council (202508330191) and PolyU Start-up Fund (P0060371). (Equal contribution: Yuxiang Luo and Qing Xu, Corresponding author: Zhen Chen)}}
        \thanks{\quad Y. Luo is with Graduate School of Information, Production and Systems, Waseda University, Japan (email: yuxiang.luo@ruri.waseda.jp).}
        \thanks{\quad Q. Xu is with University of Nottingham, UK, and with University of Nottingham Ningbo China, China (e-mail: qing.xu@nottingham.edu.cn).} 
        \thanks{\quad H. Huang is with College of Electrical Engineering and Information, Northeast Agricultural University, China (e-mail: a13220499@neau.edu.cn).}
        \thanks{\quad Y. Ouyang is with College of Computer Science, Sichuan University, China (e-mail: yuqi.ouyang@scu.edu.cn).}
        \thanks{\quad X. He is with School of Computer Science, University of Nottingham Ningbo China, China (e-mail: sean.he@nottingham.edu.cn).}
        \thanks{\quad Z. Chen is with Department of Data Science and Artificial Intelligence, The Hong Kong Polytechnic University, Hong Kong SAR. (e-mail: z.chen@polyu.edu.hk).}
        \thanks{\quad W. Duan is with School of Engineering and Physical Science, University of Lincoln, UK (email: wduan@lincoln.ac.uk).}
        \thanks{\quad J. Luo is with Department of Computer Science, University of Rochester, USA (email: jluo@cs.rochester.edu).}
        }

\markboth{IEEE Transactions on Multimedia}%
{}

\maketitle

\begin{abstract}
The 3D multi-modal brain tumor segmentation is critical to multi-modal healthcare, and it requires accurate identification of distinct internal anatomical subregions. While the recent prompt-based segmentation paradigms enable interactive experiences for clinicians, existing methods ignore cross-modal correlations and rely on labor-intensive category-specific prompts, limiting their applicability in real-world scenarios. To address these issues, we propose the MSM-Seg, a synergistic framework for multi-modal brain tumor segmentation. The MSM-Seg introduces a dual-memory segmentation paradigm that synergistically integrates multi-modal and inter-slice information with an efficient category-agnostic prompt for brain tumor understanding. To this end, we first devise a modality-and-slice memory attention (MSMA) to exploit the complex cross-modal correlations and spatial-slice dependencies among the input scans. \cz{Then, we propose a multi-scale category-agnostic prompt encoder (MCP-Encoder) to provide whole tumor region guidance for decoding.} Moreover, we devise a modality-adaptive fusion decoder (MF-Decoder) that leverages the complementary decoding information across different modalities to improve segmentation accuracy. Extensive experiments on different MRI datasets demonstrate that our MSM-Seg framework outperforms state-of-the-art methods in multi-modal metastases and glioma tumor segmentation. The code is available at \url{https://github.com/xq141839/MSM-Seg}.
\end{abstract}


\section{Introduction}
\label{sec:introduction}
The 3D multi-modal brain tumor segmentation of MRI scans is essential for precise cancer diagnosis, treatment planning, and prognosis assessment \cite{ 11329495, zhou2024shape}. This task inherently demands effective multi-modal fusion, as different MRI modalities capture fundamentally complementary tumor characteristics \cite{lei2025mlfuse, sun2025aepl}. However, the heterogeneous intensity distributions and contrast mechanisms across modalities pose significant challenges for cross-modal information integration. Furthermore, the task requires simultaneous identification of overlapping tumor components, each providing distinct clinical biomarkers for tumor grading and therapeutic decision-making \cite{liu2024causal,topff2025data}. These requirements have given rise to the challenging field of multi-modal brain tumor segmentation.

\begin{figure}[!t]
  \centering
  \includegraphics[width=1\linewidth]{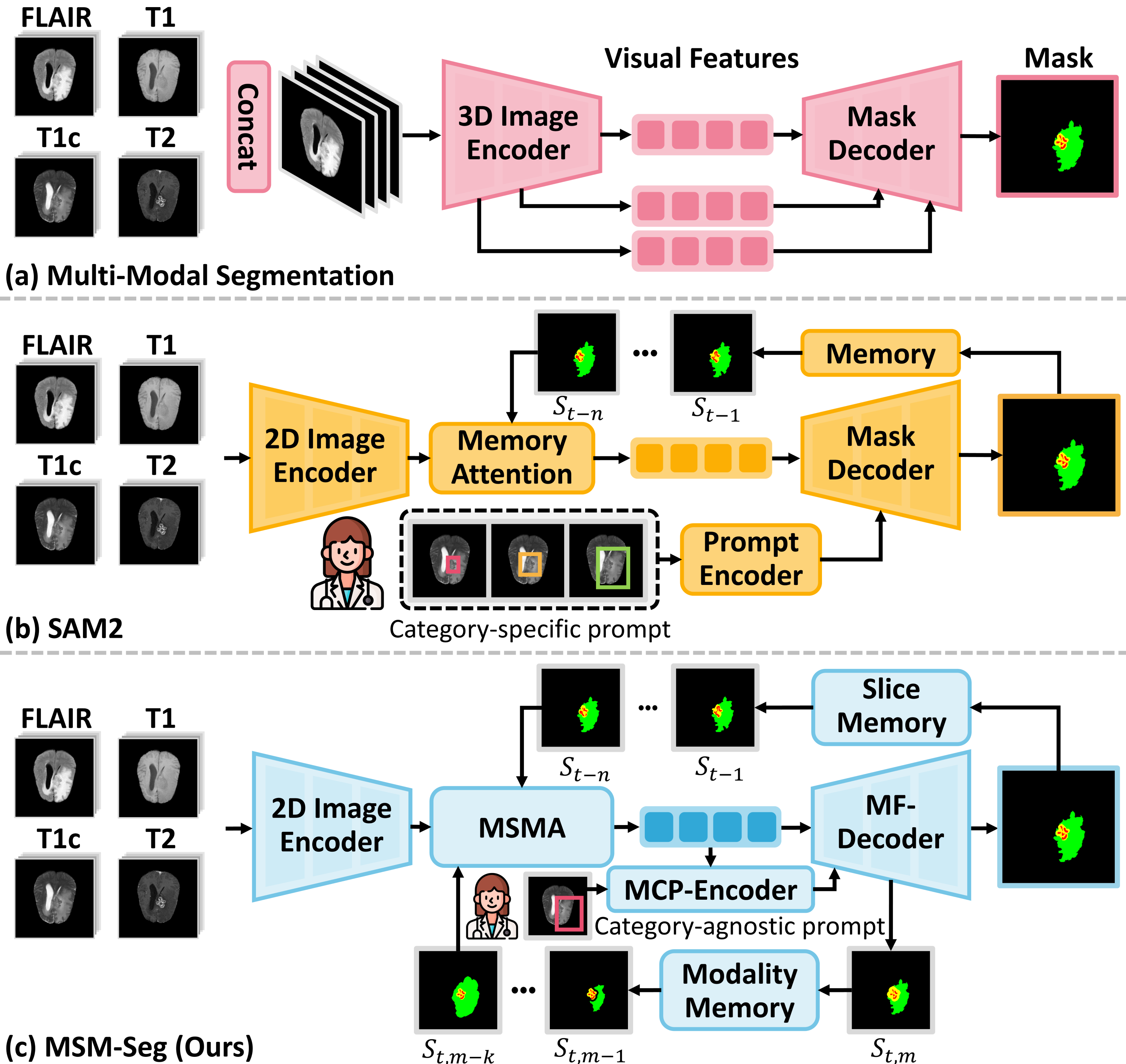}
  \caption{Comparison of our MSM-Seg and existing multi-modal brain tumor segmentation paradigms. (a) Classical multi-modal segmentation networks. (b) Recent SAM2 segmentation paradigm relies on labor-intensive and category-specific prompts. (c) MSM-Seg leverages modality and slice memory with category-agnostic prompting for multi-modal brain tumor segmentation.}
  \label{fig:intro}
\end{figure}

The classical 3D multi-modal segmentation frameworks \cite{cciccek20163d, isensee2021nnu, yang2022d, xu2023dcsau} have demonstrated considerable adaptability for diverse 3D medical imaging modalities, as illustrated in Fig. \ref{fig:intro}(a). However, existing variants \cite{xing2022nestedformer, yang20233d, shaker2024unetr++, xing2024segmamba} exhibit critical limitations from the multi-modal learning perspective. They predominantly handle multi-modal inputs through naive concatenation or independent encoding, overlooking the cross-modal correlations essential for resolving ambiguous tumor boundaries. Moreover, these frameworks treat each volume as an isolated instance, ignoring the sequential relationships between adjacent slices and the progressive evolution of tumor morphology across the scanning axis. These limitations motivate approaches that can jointly model inter-modal complementarities and intra-modal spatial continuities.

The recent segment anything model (SAM) \cite{kirillov2023segment} has revolutionized the segmentation paradigm by leveraging a large-scale vision transformer framework with interactive prompting strategies (\textit{e.g.}, bounding box and point) \cite{chen2025sam,chen2024ma,zhu2025exploiting}. In particular, SAM2 \cite{ravisam} introduced an additional memory mechanism to support medical video and 3D segmentation, as illustrated in Fig. \ref{fig:intro}(b). Despite these advancements, the adaptation of SAM2 to multi-modal brain tumor segmentation is hampered by two significant obstacles. Firstly, SAM2 relies on category-specific annotations as prompts to guide segmentation decoding. As different tumor categories exhibit overlapping spatial regions and ambiguous boundaries, such prompt modes struggle to delineate detailed subregions for each tumor type, requiring labor-intensive manual annotations that hinder clinical workflow efficiency. More notably, both current SAM2 variants \cite{zhu2024medical, yu2025crisp, yin2025memory} and previous 3D multi-modal segmentation frameworks \cite{shaker2024unetr++} typically process different MRI modalities independently or through simple concatenation, failing to exploit the rich complementary information and intrinsic cross-modal relationships. For example, FLAIR sequences excel at revealing peritumoral edema and hyperintense lesions, whereas T1c sequences provide contrast enhancement for active tumor regions, providing complementary information for tumor delineation. Therefore, their strong multi-modal complementarities motivate us to develop a unified framework that effectively captures cross-modal dependencies alongside spatial continuities for comprehensive tumor understanding.

To overcome this bottleneck, we propose the MSM-Seg for 3D multi-modal brain tumor segmentation, as illustrated in Fig. \ref{fig:intro}(c). The MSM-Seg introduces a dual-memory segmentation paradigm where different modalities, slices, and prompt information are synergistically integrated to achieve comprehensive tumor understanding. Specifically, we first devise the modality-and-slice memory attention (MSMA) that exploits cross-modal and inter-slice relationships to yield rich context-aware features. \cz{Then, we develop a multi-scale category-agnostic prompt encoder (MCP-Encoder) to generate whole tumor region guidance for the segmentation decoding process, eliminating the demand for labor-intensive category-specific annotations.} Further, we design a modality-adaptive fusion decoder (MF-Decoder) that leverages the complementary decoding information across different modalities to improve the consistency of segmentation results and reduce prediction errors, achieving superior segmentation accuracy. Experimental results on different multi-modal MRI datasets demonstrate that the MSM-Seg framework outperforms state-of-the-art methods in glioma and metastases segmentation tasks.   

The contributions of this work are summarized as follows:
\begin{itemize}
    \item We propose the MSM-Seg, a cross-modal memory-driven framework that addresses the challenge of fusing heterogeneous multi-modal information in 3D medical volumes.
    
    \item We design a MSMA to exploit the cross-modal and inter-slice relationships and enhance multi-modal feature representation.
    
    \item \cz{We devise a MCP-Encoder for whole tumor region guidance, and a MF-Decoder that leverages complementary decoding information across different modalities for precise mask generation.}
    
    \item Extensive experiments on glioma and metastases datasets demonstrate that our MSM-Seg achieves remarkable advantages over state-of-the-art multi-modal brain tumor segmentation methods.
\end{itemize}

\begin{figure*}[!t]
  \centering
  \includegraphics[width=1\linewidth]{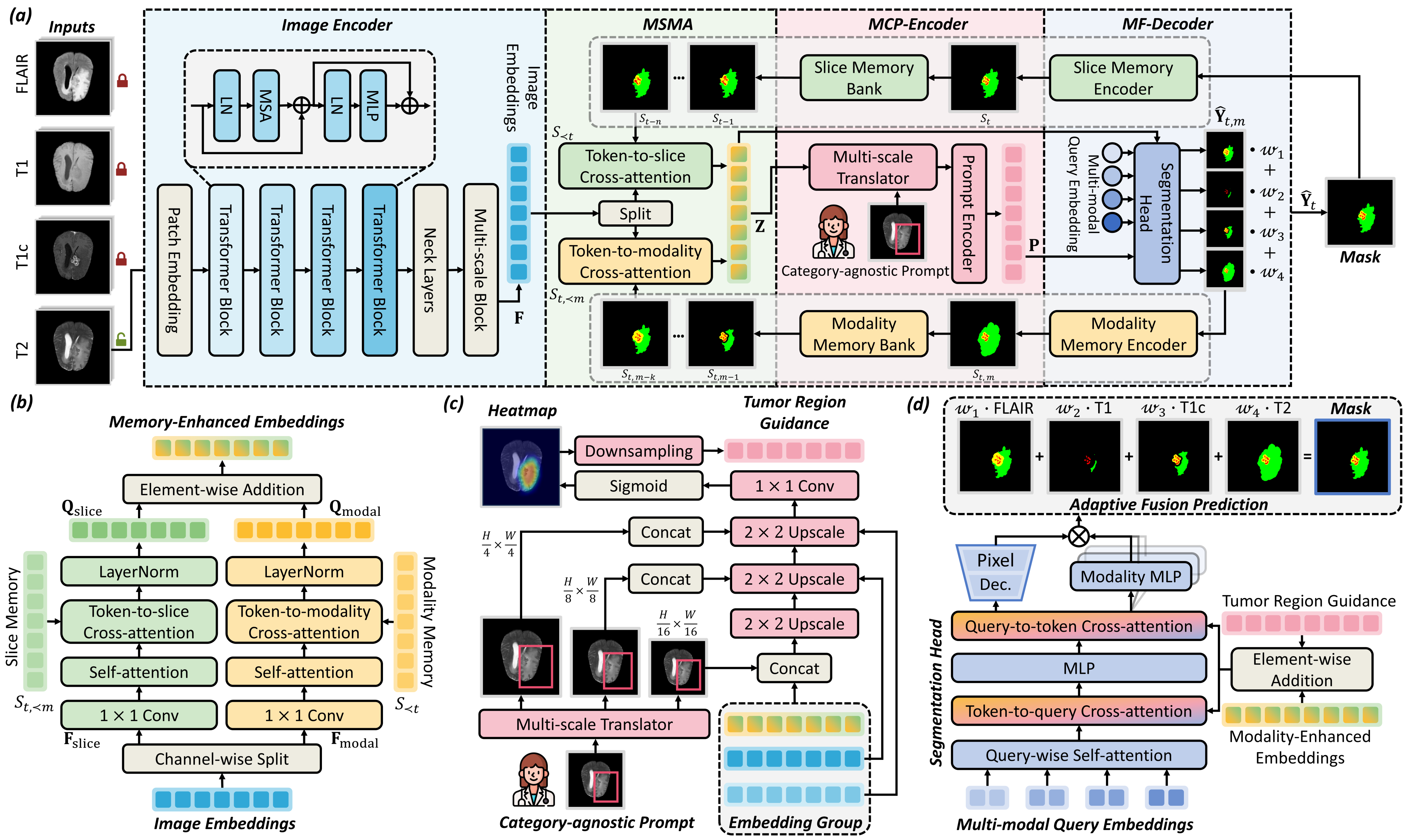}
    \caption{\cz{(a) Overview of the MSM-Seg framework for multi-modal brain tumor segmentation. At each step, MSM-Seg retrieves modality and slice memory, with (b) MSMA to create memory-enhanced embeddings, (c) MCP-Encoder to generate whole tumor region guidance using category-agnostic prompts, and (d) MF-Decoder to produce brain tumor segmentation masks. MSM-Seg effectively exploits the complementary contextual understanding conditioned on the inter-slice and cross-modal memory.}}
  \label{fig:method}
\end{figure*}

\section{Related Work}

\subsection{3D Multi-Modal Brain Tumor Segmentation}
Medical imaging technologies have undergone remarkable evolution, where MRI has emerged as the gold standard for brain tumor diagnosis and treatment planning due to its superior soft tissue contrast, non-invasive nature, and ability to provide comprehensive anatomical and functional information through multiple modality sequences such as T1, T1c, T2, and FLAIR imaging \cite{zhou2024shape,shao2025rethinking}. Each modality captures distinct tissue characteristics, offering complementary perspectives for comprehensive tumor assessment. However, the heterogeneous appearance of tumor subregions, complex anatomical boundaries, and substantial inter-patient variability in tumor morphology present significant challenges for radiologists in accurate tumor delineation and subregion identification. In this context, automated 3D multi-modal brain tumor segmentation emerges as a fundamental technology for computer-aided diagnosis, enabling precise delineation of tumor core, enhancing tumor, peritumoral edema, and necrotic regions to facilitate treatment planning and longitudinal monitoring. 

Early studies \cite{cciccek20163d, isensee2021nnu} adopted U-shaped encoder-decoder frameworks with 3D CNNs to achieve multi-modal brain tumor segmentation, typically fusing different modalities through early-stage concatenation. To enhance the capabilities of the networks in multi-modal feature encoding, recent approaches \cite{wang2021transbts, zhou2023nnformer, chen2024transunet} integrated 3D CNNs with ViT \cite{dosovitskiyimage} to capture both local spatial patterns and global contextual information of tumor subregions. Furthermore, current studies \cite{ma2024u, liu2024swin, xing2025segmamba, ye2025u} explored replacing ViT with the vision Mamba \cite{liu2024vmamba} and RWKV \cite{duanvision} to model long-range dependencies with linear computational complexity. However, these methods process volumetric data as isolated instances without efficiently leveraging inter-slice continuity, and their simplistic modality fusion strategies fail to fully exploit cross-modal complementarities inherent in multi-modal MRI data.

\subsection{Prompt-Based Segmentation}

\cz{Prompt-based segmentation has recently received increasing attention with the emergence of the segment anything model (SAM) \cite{kirillov2023segment}, which enables flexible segmentation through interactive prompts such as points and bounding boxes.} \cz{In medical image segmentation, manually prompted SAM-based methods, such as MedSAM \cite{ma2024segment} and SAMed-2 \cite{yan2025samed}, adapt SAM-style prompting to medical scenarios by using user-provided localization cues.} \cz{When applied to brain tumor segmentation, however, such prompting requires separate category-specific prompts for each tumor subregion, increasing the interaction burden.} \cz{To reduce manual interaction, automatic or self-prompting methods, such as Self-Prompt-SAM \cite{xie2025self} and AutoProSAM \cite{li2025autoprosam}, generate prompts automatically, but their generated prompts are usually more suitable for object-level or region-level localization than for stable subregion-specific guidance in multi-modal brain tumor segmentation.}

The memory mechanism has been widely applied to video object segmentation tasks where memory banks are employed to store and retrieve features from past frames to maintain temporal consistency \cite{miao2020memory}. This concept has been further revolutionized by the recent SAM2 \cite{ravisam}, which introduced a sophisticated memory bank and memory attention mechanism that enhances prediction coherence across sequential slices in volumetric scans, showing great potential for multi-modal medical image segmentation. Building upon this foundation, ReSurgSAM2 \cite{liu2025resurgsam2} incorporated additional candidate memory pools to capture diverse spatial-temporal information, while Medical SAM2 \cite{zhu2024medical} and CRISP-SAM2 \cite{yu2025crisp} optimized memory bank storage by computing similarity measures between memory slices to selectively retain the most informative temporal features. \cz{Despite these significant advances, current prompt-based and memory-based methods exhibit two critical limitations in multi-modal scenarios: they often process each MRI modality independently or through simple fusion without exploiting rich cross-modal complementarities, and they still lack an efficient category-agnostic guidance strategy tailored to nested tumor subregion segmentation.} \cz{In contrast, our MSM-Seg framework uses category-agnostic whole tumor guidance only to localize the overall tumor extent, while the fine-grained subregion differentiation is learned from multi-modal MRI features through modality-and-slice memory modeling and modality-adaptive fusion.}

\section{Methodology}
As illustrated in Fig. \ref{fig:method}(a), we present the proposed MSM-Seg framework with a dual-memory segmentation paradigm for comprehensive brain tumor understanding in MRI scans. \cz{To accomplish this, we devise the MSMA to capture cross-modal and inter-slice relationships, the MCP-Encoder to provide whole tumor region guidance for segmentation decoding, and the MF-Decoder to leverage complementary decoding information across different imaging modalities.} This integrated approach enables our MSM-Seg to leverage effective exploitation of multi-modal complementarity in MRI scans for improved brain tumor segmentation.

\subsection{Dual-Memory Segmentation Paradigm}
The classical \cite{kang2024bgf, dang2024singr} and recent SAM-based \cite{zhu2024medical, yan2025samed} multi-modal brain tumor segmentation models merge different modalities before encoding, treating them as a unified input rather than exploiting their modality-specific characteristics and cross-modal complementarities critical for accurate segmentation. To address this issue, we propose the dual-memory segmentation paradigm, allowing the MSM-Seg to perform step-wise memory integration that progressively refines its understanding of the whole tumor structure.

Given the input slice $\mathbf{X}_{t,m}$ at the step $(t, m)$, the model maintains a latent state $\mathcal{S}_{t,m} \in \mathbb{R}^{C \times H \times W}$, where $t \in \{1, \dots, T\}$ is the slice index, and $m \in \{1, \dots, M\}$ is the modality index for input slice $\mathbf{X}_{t,m}$. With $\mathcal{S}_{t,m}$ representing the current contextual understanding conditioned on previous slices and modalities, MSM-Seg receives an input slice $\mathbf{X}_{t,m}$ and updates its internal state as follows:
\begin{equation}
\begin{cases}
    \mathcal{S}_{t,m} = \mathcal{R}(\mathbf{X}_{t,m}, \theta_{t, m}, \mathcal{S}_{t,\prec m}, \mathcal{S}_{\prec t}), \\
    \hat{\mathbf{Y}}_{t,m} = \mathcal{P}(\mathcal{S}_{t,m}).
\end{cases}
\end{equation}

Here:
\begin{itemize}
    \item $\mathcal{R}(\cdot)$ is the state update function for integrating multi-modal, inter-slice, and prompt information.
    \item $\mathcal{P}(\cdot)$ is the segmentation prediction head that maps the state to the predicted mask.
    \item $\mathcal{S}_{t,\prec m}$ denotes the cross-modal context aggregated from previous modalities $m-k$ to $m-1$ at current slice $t$.
    \item $\mathcal{S}_{\prec t}$ denotes the inter-slice context aggregated from the previous slices $t-n$ to $t-1$.
    \item \cz{$\theta_{t, m}$ denotes the efficient category-agnostic prompt used to generate whole tumor region guidance.}
\end{itemize}

After the model has processed all modalities $\{1, \dots, M\}$ for a given slice $t$, we obtain a set of modality-specific predictions $\{ \hat{\mathbf{Y}}_{t,1}, \dots, \hat{\mathbf{Y}}_{t,M} \}$. These are combined to produce the final fused segmentation result $\hat{\mathbf{Y}}_t$ as follows:
\begin{equation}
\hat{\mathbf{Y}}_t = \mathcal{F}(\{\hat{\mathbf{Y}}_{t,m}\}_{m=1}^M),
\end{equation}
where $\mathcal{F}(\cdot)$ performs the adaptive weighting of modality-specific outputs, enabling the model to dynamically select the most informative modalities at each voxel. To capture contextual information $\mathcal{S}_{t,\prec m}$ and $\mathcal{S}_{\prec t}$, we construct two memory banks, as follows:

\begin{itemize}
    \item \textbf{Modality Memory Bank:} For each modality of current slice $(t, m)$, its prediction $\hat{\mathbf{Y}}_{t,m}$ is encoded via a modality memory encoder: $\mathbf{V}_{t,m} = f_{\rm modal}(\hat{\mathbf{Y}}_{t,m})$. These are aggregated to produce the cross-modal context: $\mathcal{S}_{\prec t,m} = \{\mathbf{V}_{t,m}\}_{m=1}^k$, where $k < M$.

    \item \textbf{Slice Memory Bank:} For each slice $t$, its segmentation prediction $\hat{\mathbf{Y}}_{t}$ is encoded using a slice memory encoder $\mathbf{U}_{t} = f_{\rm slice}(\hat{\mathbf{Y}}_{t})$. These memory vectors are then aggregated to yield the inter-slice context $\mathcal{S}_{\prec t} = \{\mathbf{U}_{t}\}_{t=1}^n$.
    
\end{itemize}
\cz{In this way, the dual-memory segmentation paradigm breaks the isolation between modalities and slices through externalized memory banks, enabling MSM-Seg to achieve synergistic optimization by integrating inter-slice and cross-modal information with whole tumor region guidance.}

\subsection{Modality-and-Slice Memory Attention}
The dual-memory segmentation paradigm is proposed to establish contextual understanding conditioned on previous slices and modalities. To achieve this, we introduce the MSMA to efficiently exploit the cross-modal and inter-slice relationships, as shown in Fig. \ref{fig:method}(b). Given the image embedding $\mathbf{F} \in \mathbb{R}^{C \times H \times W}$ extracted from the input slice $\mathbf{X}_{t,m}$, we split it evenly along the channel dimension, as follows:
\begin{equation} \label{eq:3}
[\mathbf{F}_{\text{slice}}, \mathbf{F}_{\text{modal}}] = \text{Split}(\mathbf{F}),
\end{equation}
where $\mathbf{F}_{\text{slice}}$ and $ \mathbf{F}_{\text{modal}}$ are used to handle slice memory and modality memory, respectively. To align the separated image embedding with each memory, we adopt two $1 \times 1$ convolutions $\phi(\cdot)$ for up-projection. Then, we utilize self-attention $\mathcal{SA}(\cdot)$ to update both embeddings. The above operations can be formulated as follows:
\begin{equation} \label{eq:4}
\mathbf{Q}_{\text{slice}} = \mathcal{SA}(\phi(\mathbf{F}_{\text{slice}})), \quad \mathbf{Q}_{\text{modal}} = \mathcal{SA}(\phi(\mathbf{F}_{\text{modal}})).
\end{equation}
Furthermore, we apply cross-attention, denoted as $\mathcal{CA}(\cdot)$ to integrate contextual information from memory banks. The queries ($\mathbf{Q}_{\text{slice}}, \mathbf{Q}_{\text{modal}}$) attend to their corresponding key-value pairs drawn from slice memory $\mathcal{S}_{\prec t}$, and modality memory $\mathcal{S}_{\prec t,m}$. The final memory-enhanced embedding is computed as follows:
\begin{equation} \label{eq:5}
\begin{split}
\mathbf{Z} &=\mathcal{CA}(Q=\mathbf{Q}_{\text{slice}}, K=V=\mathcal{S}_{\prec t}, \mathcal{S}_{\prec t}) \\ &+ \mathcal{CA}(Q=\mathbf{Q}_{\text{modal}}, K=V=\mathcal{S}_{\prec t,m}, \mathcal{S}_{\prec t,m}).
\end{split}
\end{equation}
\cz{On this basis, we apply the proposed MSMA to effectively capture the cross-modal and inter-slice relationships, producing memory-enhanced representations that serve as the foundation for subsequent whole tumor region guidance and segmentation mask decoding.}

\begin{algorithm}[t]
\caption{Optimization Pipeline for MSM-Seg}
\label{alg:msmseg}
\KwIn{
MRI scan $\{\mathbf{X}_{t,m}\}_{t=1,m=1}^{T,M}$ with $T$ slices and $M$ modalities;\\
\hspace{3.0em}Optional user prompt $\mathcal{Q}_{\text{box}}$;\\
\hspace{3.0em}Pretrained image encoder $E$;\\
}
\KwOut{Final segmentation masks $\{\hat{\mathbf{Y}}_t\}_{t=1}^T$}

\For{slice $t = 1$ to $T$}{
  \For{modality $m = 1$ to $M$}{
    Extract feature: $\mathbf{F}_{t,m} = E(\mathbf{X}_{t,m})$\;
    
    Retrieve memory: $\mathcal{S}_{\prec t}$ and $\mathcal{S}_{t,\prec m}$ (if exist)\;

    Compute: $\mathbf{Z}_{t,m} = \text{MSMA}(\mathbf{F}_{t,m}, \mathcal{S}_{\prec t}, \mathcal{S}_{t,\prec m})$ via Eq. \eqref{eq:3}-\eqref{eq:5}\;

    \cz{Generate whole tumor region guidance}: $\mathbf{P}_{t,m} = \text{MCP-Encoder}(\mathbf{Z}_{t,m},\mathcal{Q}_{\text{box}},\{\mathbf{F}_i\}_{i=1}^{l-1})$\ via Eq. \eqref{eq:6}-\eqref{eq:7}\;

    Decode features: $\hat{\mathbf{Y}}_{t,m} = \text{MF-Decoder}(\mathbf{H}_{t,m},\mathbf{P}_{t,m},\mathbf{E}_{t,m})$ using $\mathbf{H}_{t,m} = \mathbf{Z}_{t,m} \oplus \mathbf{P}_{t,m}$ and Eq. \eqref{eq:8}\;

    \tcp{Update modality memory}
    Encode and store: $\mathbf{V}_{t,m} = f_{\text{modal}}(\hat{\mathbf{Y}}_{t,m})$\;
  }
  Fuse predictions $\hat{\mathbf{Y}}_t$ using Eq. \eqref{eq:9}\;

  \tcp{Update slice memory}
  Encode and store: $\mathbf{U}_t = f_{\text{slice}}(\hat{\mathbf{Y}}_t)$\;
}
\end{algorithm}

\subsection{Multi-Scale Category-Agnostic Prompt Encoder}
Existing prompt-based segmentation methods \cite{ravisam, zhu2024medical, yan2025samed} typically require category-specific prompts for each tumor subregion (\textit{e.g.}, tumor core, enhancing tumor, peritumoral edema), which necessitate labor-intensive manual annotation and prior anatomical knowledge. This category-specific paradigm not only increases clinical workload but also limits the model's flexibility in handling diverse tumor presentations. \cz{To address this issue, we propose the MCP-Encoder to provide whole tumor region guidance, as illustrated in Fig. \ref{fig:method}(c).} Specifically, our MCP-Encoder operates in two modes:
\begin{itemize}
    \item \textbf{Category-Agnostic Prompt Mode:} We incorporate category-agnostic prompts (\textit{e.g.}, bounding boxes) from users for the generation of potential tumor regions. Unlike existing SAM series \cite{kirillov2023segment, ravisam, yan2025samed} that require category-specific annotations for each tumor subregion, our approach only needs a single bounding box covering the entire tumor region. This category-agnostic design significantly reduces annotation burden and enhances clinical applicability.
    
    \item \textbf{Automatic Mode:} \cz{Without any manual annotations, whole tumor region guidance is autonomously generated by leveraging sufficient feature representations from the image encoder and our MSMA. This mode enables fully self-guided segmentation across diverse tumor regions.}
 
\end{itemize} 
Moreover, brain tumors vary significantly in scale, from small satellite lesions to large infiltrative masses. Therefore, our MCP-Encoder receives a set of multi-scale features \( \{ \mathbf{F}_i \}_{i=1}^{l-1} \) from the shared backbone, concatenating with the memory-enhanced representation $\mathbf{Z}$ to establish an embedding group, where $i\in\{1,\cdots,l\}$ denotes the number of layers in the image encoder. In our category-agnostic prompt mode, users' prompt input \( \{ \mathbf{G}_i \}_{i=1}^{l} \) is encoded through a multi-scale translator, which employs bilinear interpolation to resize users' prompt input for matching each feature scale. A layer-by-layer multi-scale fusion is then performed:
\begin{equation} \label{eq:6}
\mathbf{F}_i^{\text{fusion}} =
\begin{cases}
\text{Concat}(\mathbf{F}_{i-1}^{\text{fusion}}, \mathbf{F}_i, \mathbf{G}_i), & \text{if prompt available,} \\
\text{Concat}(\mathbf{F}_{i-1}^{\text{fusion}}, \mathbf{F}_i), & \text{otherwise.}
\end{cases}
\end{equation}
\cz{The final whole tumor region guidance can be generated as follows:}
\begin{equation} \label{eq:7}
\mathbf{P} = {\mathcal{DS}}(\sigma(\phi(\mathbf{F}_{l}^{\text{fusion}}))),
\end{equation}
where $\mathcal{DS}(\cdot)$ stands for two $3 \times 3$ convolutions followed by one $1 \times 1$ convolution for downsampling, $\sigma(\cdot)$ represents an activation function, \textit{i.e.}, sigmoid. \cz{In this way, the proposed MCP-Encoder removes the requirement for labor-intensive category-specific prompts through its category-agnostic design, enabling MSM-Seg to provide efficient whole tumor region guidance that captures multi-scale tumor characteristics while significantly reducing the clinical annotation burden.}

\subsection{Modality-Adaptive Fusion Decoder}
In the field of multi-modal brain tumor segmentation, each MRI modality captures unique tumor characteristics, \textit{e.g.}, T1c highlights contrast-enhancing cores. However, existing methods employ static fusion strategies that fail to exploit such modality-specific strengths adaptively. To this end, we devise the MF-Decoder on the basis of the dual-memory segmentation paradigm to further leverage the distinct complementary information across different modalities for accurate tumor segmentation. \cz{As shown in Fig. \ref{fig:method}(d), for each modality \( m \) at slice \( t \), our MF-Decoder receives the memory-enhanced embedding \( \mathbf{Z}_{t,m} \) and its corresponding whole tumor region guidance \( \mathbf{P}_{t,m} \).} These are fused with a multi-modal query embedding $
\mathbf{E}_{t,m}$ through a stack of attention layers, including query-wise self-attention, token-to-query and query-to-token cross-attention, and MLP transformations. The prompt embedding is integrated via element-wise addition to the memory-enhanced image embedding: $\mathbf{H}_{t,m} = \mathbf{Z}_{t,m} \oplus \mathbf{P}_{t,m}$. Each refined image embedding \( \mathbf{H}_{t,m} \) is passed through a modality segmentation head consisting of a shared pixel-wise decoder $\mathcal{P}_{\rm pd}(\cdot)$ and a modality-specific MLP $\mathcal{P}_{\rm mlp}(\cdot)$, and produces the modality-specific prediction:
\begin{equation} \label{eq:8}
\hat{\mathbf{Y}}_{t,m} = \mathcal{P}_{\rm pd}(\mathbf{H}_{t,m}) \otimes \mathcal{P}_{\rm mlp}(\mathbf{E}_{t,m}),
\end{equation}
\cz{where $\otimes$ denotes element-wise multiplication. To obtain the final segmentation mask \( \hat{\mathbf{Y}}_t \), we apply an adaptive weighting strategy as follows:}
\begin{equation} \label{eq:9}
\hat{\mathbf{Y}}_t = \sum_{m=1}^{M} w_{m} \cdot \hat{\mathbf{Y}}_{t,m}.
\end{equation}
In this way, the proposed MF-Decoder breaks the limitation of uniform modality treatment through adaptive fusion, enabling our MSM-Seg framework to achieve synergistic multi-modal segmentation by fully exploiting the complementary strengths of different imaging modalities in MRI scans.

\begin{table*}[!t]
\centering
\small
\caption{\cz{Comparison with state-of-the-arts on brain tumor segmentation. The MP represents the manual prompt.}}
\setlength\tabcolsep{2.7pt}
\scalebox{1}{\begin{tabular}{l|c|cccc|cccc|cccc|cccc}
\hline
&  & \multicolumn{8}{c|}{BraTS-METS} & \multicolumn{8}{c}{BraTS-AGPT} \\
\cline{3-18}
Methods & MP & \multicolumn{4}{c|}{Dice (\%) $\uparrow$} & \multicolumn{4}{c|}{HD95 (mm) $\downarrow$} & \multicolumn{4}{c|}{Dice (\%) $\uparrow$} & \multicolumn{4}{c}{HD95 (mm) $\downarrow$} \\
\cline{3-18}
& & ET & TC & WT & Avg & ET & TC & WT & Avg & ET & TC & WT & Avg & ET & TC & WT & Avg \\
\hline
TransBTS \cite{wang2021transbts} & \multirow{9}{*}{\usym{2715}} & 63.68 & 63.55 & 67.00 & 66.28 & 25.22 & 24.99 & 25.92 & 25.45 & 58.42 & 60.33 & 81.75 & 70.17 & 12.85 & 11.47 & 10.92 & 11.75 \\
EoFormer \cite{she2023eoformer} & & 67.39 & 67.29 & 70.63 & 69.97 & 26.70 & 27.01 & 28.57 & 27.79 & 61.73 & 62.56 & 83.29 & 72.53 & 11.94 & 10.83 & 10.15 & 10.97 \\
3D-TransUNet \cite{yang20233d} & & 65.84 & 64.77 & 69.92 & 68.18 & 24.31 & 23.88 & 24.73 & 24.31 & 60.29 & 62.41 & 82.67 & 71.46 & 12.43 & 11.29 & 10.58 & 11.43 \\
SiNGR \cite{dang2024singr} & & 66.53 & 70.14 & 71.88 & 69.52 & 23.89 & 23.42 & 24.06 & 23.79 & 62.18 & 64.73 & 84.15 & 73.35 & 11.67 & 10.45 & 9.93 & 10.68 \\
UNETR++ \cite{shaker2024unetr++} & & 67.82 & 71.95 & 73.64 & 71.14 & 22.47 & 21.83 & 22.37 & 22.22 & 63.75 & 65.94 & 85.42 & 74.70 & 10.89 & 9.78 & 9.24 & 9.97 \\
nnUnet-V2 \cite{isensee2024nnu} & & \textbf{68.94} & 72.87 & 74.16 & 73.87 & \textbf{18.32} & 20.65 & 20.91 & 20.82 & 64.87 & 67.58 & 85.21 & 75.12 & 9.73 & 8.91 & 8.67 & 9.04 \\
SegMamba-V2 \cite{xing2025segmamba} & & 68.41 & \underline{73.08} & \underline{74.77} & \underline{73.92} & 19.52 & \underline{20.14} & \underline{20.65} & \underline{20.40} & 65.24 & 68.67 & 85.03 & 76.49 & \underline{6.52} & \underline{6.89} & 8.87 & 8.15 \\
\cz{SuperLightNet \cite{yu2025superlightnet}} & & \cz{67.82} & \cz{72.15} & \cz{74.03} & \cz{73.28} & \cz{20.13} & \cz{21.04} & \cz{21.46} & \cz{21.08} & \cz{\underline{65.27}} & \cz{\underline{69.04}} & \cz{\underline{85.62}} & \cz{\underline{77.16}} & \cz{7.47} & \cz{7.83} & \cz{\underline{8.41}} & \cz{\underline{7.90}} \\
\cz{MSM-Seg (Ours)} & & \underline{68.53} & \textbf{74.11} & \textbf{75.89} & \textbf{75.01} & \underline{18.57} & \textbf{19.19} & \textbf{19.40} & \textbf{19.29} & \textbf{65.61} & \textbf{70.32} & \textbf{87.10} & \textbf{78.71} & \textbf{6.17} & \textbf{6.74} & \textbf{7.29} & \textbf{7.02} \\
\hline
SAM \cite{kirillov2023segment} & \multirow{8}{*}{\usym{1F5F8}} & 66.84 & 72.23 & 73.42 & 72.83 & 18.74 & 17.92 & 16.31 & 17.66 & 61.29 & 64.57 & 84.83 & 75.90 & 8.15 & 7.41 & 6.73 & 7.43 \\
MA-SAM \cite{chen2024ma} & & 67.91 & 73.85 & 74.68 & 74.15 & 17.83 & 16.97 & 15.74 & 16.85 & 62.74 & 65.94 & 86.29 & 77.32 & 7.82 & 7.09 & 6.42 & 7.11 \\
SAM2 \cite{ravisam} & & 68.73 & 74.94 & 75.87 & 75.18 & 16.92 & 16.15 & 14.87 & 15.98 & 63.85 & 68.17 & 87.64 & 78.55 & 7.24 & 6.83 & 6.15 & 6.74 \\
MedSAM-2 \cite{zhu2024medical} & & 69.47 & 75.81 & 76.92 & 76.07 & 16.24 & 15.49 & 14.22 & 15.32 & 64.92 & 70.73 & 88.95 & 79.87 & 6.91 & 6.47 & 5.89 & 6.42 \\
SAM2-Adapter \cite{chen2025sam2} & & 69.92 & 76.64 & 77.83 & 76.80 & 15.67 & 14.91 & 13.68 & 14.75 & 65.18 & 69.19 & 89.27 & 80.96 & 6.63 & 6.21 & 5.67 & 6.17 \\
SAMed-2 \cite{yan2025samed} & & 70.15 & 77.21 & \underline{79.04} & \underline{77.47} & \underline{15.24} & \underline{14.35} & \underline{13.21} & \underline{14.27} & 65.47 & 71.41 & 89.73 & 81.44 & \underline{6.38} & \underline{5.97} & \underline{5.43} & \underline{6.12} \\
\cz{SAM-Med3D \cite{wang2025sam}} & & \cz{\underline{70.43}} & \cz{\underline{77.65}} & \cz{78.32} & \cz{77.21} & \cz{15.82} & \cz{14.93} & \cz{13.76} & \cz{14.83} & \cz{\underline{66.14}} & \cz{\underline{72.43}} & \cz{\underline{90.31}} & \cz{\underline{82.48}} & \cz{6.63} & \cz{5.98} & \cz{5.51} & \cz{6.15} \\
\cz{MSM-Seg (Ours)} & & \textbf{71.82} & \textbf{79.45} & \textbf{81.27} & \textbf{79.51} & \textbf{14.67} & \textbf{13.84} & \textbf{12.73} & \textbf{13.75} & \textbf{67.29} & \textbf{73.76} & \textbf{92.43} & \textbf{83.84} & \textbf{5.94} & \textbf{5.21} & \textbf{5.15} & \textbf{5.56} \\
\hline
\end{tabular}}
\label{tab:compar}
\end{table*}

\begin{table*}[!t]
    \centering
    \small
    \caption{\cz{Ablation study of MSM-Seg on BraTS-METS and BraTS-AGPT in the category-agnostic prompt mode using whole tumor bounding box prompts. $A$: MSMA, $E$: MCP-Encoder, $D$: MF-Decoder, $P$: dual-memory segmentation paradigm.}}
    \setlength\tabcolsep{4.5pt}
    \scalebox{0.97}{\begin{tabular}{ccccc|cccc|cccc|cccc|cccc}
    \hline
    & \multirow{3}{*}{$\mathcal{A}$} & \multirow{3}{*}{$\mathcal{E}$} & \multirow{3}{*}{$\mathcal{D}$} & \multirow{3}{*}{$\mathcal{P}$} & \multicolumn{8}{c|}{BraTS-METS} & \multicolumn{8}{c}{BraTS-AGPT} \\
    \cline{6-21}
    & & & & & \multicolumn{4}{c|}{Dice (\%) $\uparrow$} & \multicolumn{4}{c|}{HD95 (mm) $\downarrow$} & \multicolumn{4}{c|}{Dice (\%) $\uparrow$} & \multicolumn{4}{c}{HD95 (mm) $\downarrow$}\\
    \cline{6-21}
    & & & & & ET & TC & WT & Avg & ET & TC & WT & Avg & ET & TC & WT & Avg & ET & TC & WT & Avg\\
    \hline
    1 & &  &  &  & 68.73 & 74.94 & 75.87 & 75.18 & 16.92 & 16.15 & 14.87 & 15.98 & 63.85 & 68.17 & 87.64 & 78.55 & 7.24 & 6.83 & 6.15 & 6.74\\ 
    2 & \usym{1F5F8} &  &  &  & 69.21 & 75.68 & 76.75 & 75.83 & 16.56 & 15.78 & 14.52 & 15.62 & 64.39 & 69.05 & 89.32 & 79.36 & 7.03 & 6.58 & 5.97 & 6.55\\
    3 & \usym{1F5F8} & \usym{1F5F8} & &  & 69.85 & 76.58 & 78.15 & 76.70 & 16.08 & 15.28 & 14.05 & 15.14 & 65.12 & 70.38 & 90.72 & 80.43 & 6.76 & 6.24 & 5.53 & 6.30\\
    4 & \usym{1F5F8} & \usym{1F5F8} & \usym{1F5F8} &  & 70.42 & 77.35 & 79.18 & 77.78 & 15.75 & 14.89 & 13.68 & 14.77 & 65.74 & 71.48 & 90.63 & 81.76 & 6.53 & 5.98 & 5.55 & 6.00\\
    5 & \usym{1F5F8} & \usym{1F5F8} & \usym{1F5F8} & \usym{1F5F8} & \textbf{71.82} & \textbf{79.45} & \textbf{81.27} & \textbf{79.51} & \textbf{14.67} & \textbf{13.84} & \textbf{12.73} & \textbf{13.75} & \textbf{67.29} & \textbf{73.76} & \textbf{92.43} & \textbf{83.84} & \textbf{5.94} & \textbf{5.21} & \textbf{5.15} & \textbf{5.56}\\
    \hline
    \end{tabular}}
    \label{tab:ab}
\end{table*}

\subsection{Training and Optimization Pipeline}

We summarize the training and optimization pipeline of MSM-Seg for multi-modal brain tumor segmentation in Algorithm~\ref{alg:msmseg}. To construct our MSM-Seg framework, we first initialize the image encoder ($E$), MSMA module ($\mathcal{A}$), MCP-Encoder ($\mathcal{E}$), and MF-Decoder ($\mathcal{D}$), ensuring the consistency of feature learning. The framework follows a nested training strategy with an outer loop iterating across slices ($t = 1$ to $T$) and an inner loop processing modalities sequentially ($m = 1$ to $M$) within each slice, which enables progressive integration of both inter-slice and cross-modality context. 

For each modality $m$ in slice $t$, the image encoder $E$ first extracts features $\mathbf{F}_{t,m}$ from the input MRI scan $\mathbf{X}_{t,m}$. The MSMA module then retrieves relevant memories from previous slices ($\mathcal{S}_{<t}$) and previous modalities within the current slice ($\mathcal{S}_{t,<m}$) to compute memory-enhanced embeddings $\mathbf{Z}_{t,m}$. \cz{Subsequently, the MCP-Encoder generates whole tumor region guidance $\mathbf{P}_{t,m}$ by incorporating the enhanced embeddings with optional category-agnostic prompts $\mathcal{Q}_{\text{box}}$.} The MF-Decoder produces modality-specific segmentation masks $\hat{\mathbf{Y}}_{t,m}$ using the concatenated features $\mathbf{H}_{t,m}$ and tumor guidance. After processing all modalities within slice $t$, the framework fuses individual predictions via adaptive fusion to generate the final slice segmentation $\hat{\mathbf{Y}}_{t}$. 

Throughout training, both modality-specific and slice-level memories are continuously updated and stored in their respective memory banks, which are constructed following \cite{ravisam}. \cz{Moreover, the training of MSM-Seg consists of three parts: (1) the optimization of whole tumor region guidance $\mathbf{P}$, (2) the joint optimization of each modality-specific segmentation mask $\hat{\mathbf{Y}}_{t,m}$, and (3) the optimization of the final brain tumor segmentation mask $\hat{\mathbf{Y}}_{t}$.} The overall loss function is formulated as follows:
\begin{equation}
    \mathcal{L}_{\rm MSMSeg} = \mathcal{L}_{\rm prompt} + \frac{1}{M}\sum_{m=1}^{M}\mathcal{L}^m_{\rm modal} + \mathcal{L}_{\rm fusion},
\end{equation}
where $\mathcal{L}_{\rm prompt}$ adopts the binary cross-entropy loss, while $\mathcal{L}^m_{\rm modal}$ and $\mathcal{L}_{\rm fusion}$ employ the combination of dice and focal loss. By jointly optimizing $\mathcal{L}_{\rm MSMSeg}$, MSM-Seg achieves accurate multi-modal brain tumor segmentation with superior performance across diverse subregions.

\section{Experiments}

\subsection{Datasets and Implementations}
\subsubsection{Datasets}
To validate the effectiveness of the proposed MSM-Seg framework, we conduct comprehensive evaluations on two multi-modal brain tumor segmentation datasets from the BraTS 2024 challenge, as follows:

\noindent \textbf{BraTS-METS} \cite{de20242024} is a brain metastases segmentation dataset that addresses the segmentation of secondary brain tumors originating from primary cancers. The dataset includes 652 cases with multi-contrast MRI examinations.

\noindent \textbf{BraTS-AGPT} \cite{moawad2024brain} is an adult post-treatment glioma segmentation dataset that focuses on segmenting residual or recurrent gliomas following therapeutic intervention. The dataset contains 1,349 cases with four imaging modalities.

\subsubsection{Implementation Details}
We perform all experiments on one NVIDIA A6000 GPU with PyTorch. Following \cite{zhu2024medical, yan2025samed}, we adopt the Hirea-S structure \cite{ryali2023hiera} as the image encoder of our MSM-Seg and initialize it with the pretrained SAM2 weight \cite{chen2025sam2}. We apply the optimizer using AdamW ($\beta_1$=0.9, $\beta_2$=0.999) with an initial learning rate of $1\times10^{-4}$, a weight decay of 0.01, and use the exponential decay strategy to adjust the learning rate with a factor of 0.98. The batch size and the training epoch are set as 16 and 300. For fair comparisons, all methods are implemented with the same training setting. We follow a standard random data split ratio of 7:1:2, and all images are resized to $256 \times 256$ during training and evaluation stages. \cz{As each case in both datasets provides four different modalities, we set the modality memory bank $k$ as 3 and follow \cite{chen2025sam2, zhu2024medical, yan2025samed} to set the slice memory bank $n$ as 8.} For evaluating the upper-bound performance of all SAM methods, we apply bounding boxes as prompts following \cite{ma2024segment, yan2025samed}.

\begin{table*}[!t]
    \centering
    \small
    \caption{\cz{Ablation study of modality sequence on BraTS-METS and BraTS-AGPT in the category-agnostic prompt mode. Our MSM-Seg uses the default sequence of FLAIR→T1→T1c→T2. The t-test analysis of Dice results with P-value $>$ 0.05 demonstrates that the performance of MSM-Seg is not significantly affected by the input sequence of MRI modalities.}}
    \setlength\tabcolsep{2.6pt}
    \scalebox{1}{\begin{tabular}{c|cccc|cccc|cccc|cccc|c}
    \hline
    \multirow{3}{*}{Modality Sequence} & \multicolumn{8}{c|}{BraTS-METS} & \multicolumn{8}{c|}{BraTS-AGPT} & \multirow{3}{*}{P-value}\\
    \cline{2-17}
    & \multicolumn{4}{c|}{Dice (\%) $\uparrow$} & \multicolumn{4}{c|}{HD95 (mm) $\downarrow$} & \multicolumn{4}{c|}{Dice (\%) $\uparrow$} & \multicolumn{4}{c|}{HD95 (mm) $\downarrow$} & \\
    \cline{2-17}
     & ET & TC & WT & Avg & ET & TC & WT & Avg & ET & TC & WT & Avg & ET & TC & WT & Avg &\\
    \hline
    T2→FLAIR→T1→T1c & 71.76 & {79.63} & 81.39 & 79.56 & 14.79 & 13.72 & 12.61 & 13.81 & 67.22 & 73.69 & 92.56 & 83.72 & 6.01 & 5.25 & {5.05} & 5.58 & 0.09\\
    T1c→T2→FLAIR→T1 & 71.69 & 79.31 & 81.18 & 79.43 & 14.73 & 13.95 & 12.82 & 13.89 & 67.25 & 73.58 & 92.29 & 83.74 & 5.97 & 5.28 & 5.21 & 5.62 & 0.12\\
    T1→T1c→T2→FLAIR & 71.78 & 79.61 & {81.42} & 79.48 & 14.75 & {13.69} & {12.58} & 13.77 & 67.23 & 73.73 & {92.59} & 83.81 & 6.03 & {5.13} & 5.07 & 5.59 & 0.15\\
    \hline
    FLAIR→T1→T1c→T2 & {71.82} & 79.45 & 81.27 & {79.51} & {14.67} & 13.84 & 12.73 & {13.75} & {67.29} & {73.76} & 92.43 & {83.84} & {5.94} & 5.21 & 5.15 & {5.56} & -\\
    \hline
    \end{tabular}}
    \label{tab:modality_sequence}
\end{table*}

\begin{figure*}[!t]
\centering
  \includegraphics[width=1\linewidth]{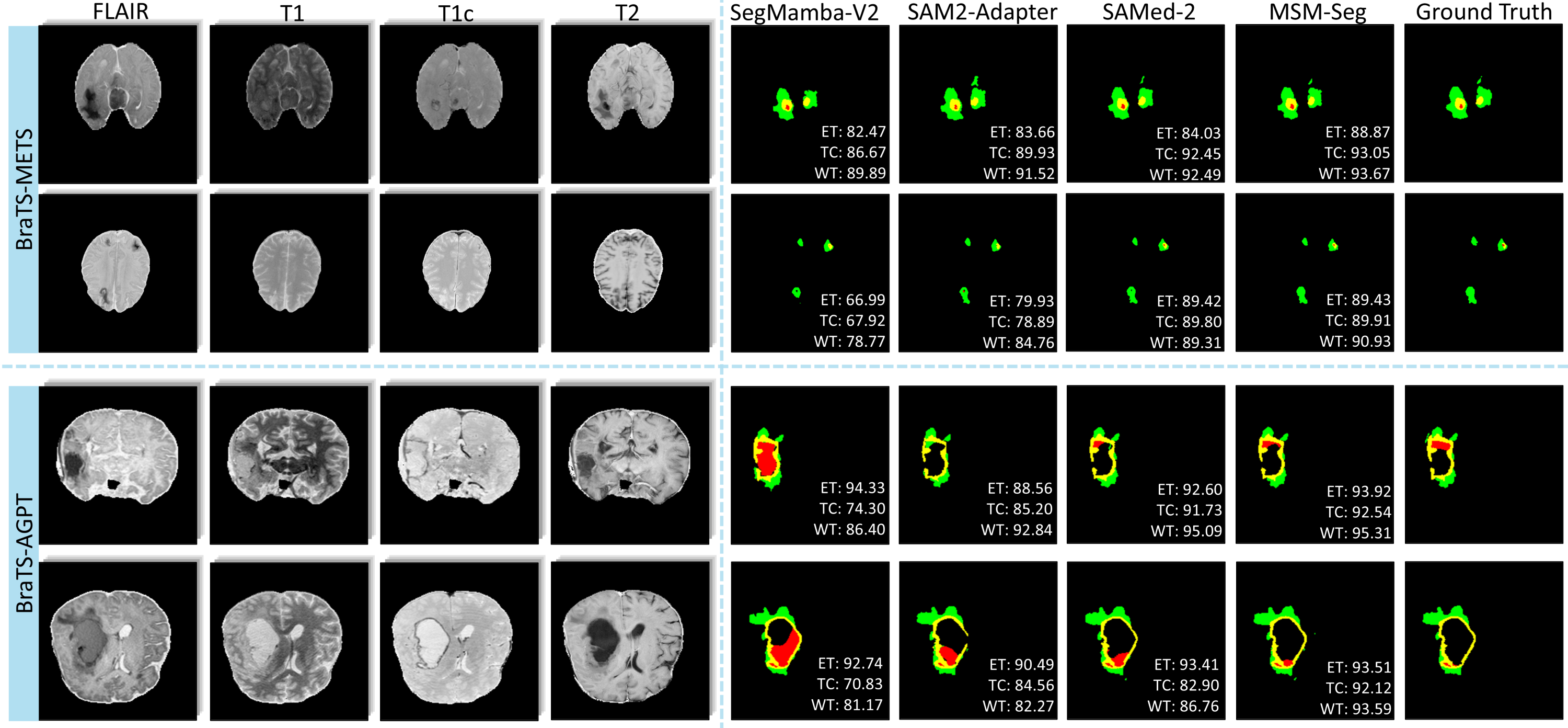}
  \caption{Visualization of multi-modal brain tumor segmentation on the BraTS-METS and BraTS-AGPT datasets. Our MSM-Seg exhibits the best results, accurately delineating hierarchical tumor subregions (\textit{i.e.}, ET, NETC, and SNFH) with precise boundaries while having fewer false positives and better preservation of tumor morphology compared to state-of-the-art methods.}
  \label{fig:visual1}
\end{figure*}

\subsubsection{Evaluation Metrics}
To perform a comprehensive evaluation of multi-modal brain tumor segmentation, we adopt standard metrics following the BraTS challenge protocol. Specifically, we evaluate the segmentation performance on three hierarchical tumor regions with nested structures: enhancing tumor (ET), tumor core (TC), and whole tumor (WT). The ET represents the enhancing tumor regions. The TC is computed as the union of ET and surrounding non-enhancing FLAIR hyperintensity (SNFH). The WT is calculated as the union of TC and non-enhancing tumor core (NETC), defining the whole extent of the tumor \cite{maleki2025analysis}. For quantitative assessment, we employ the Dice similarity coefficient and 95\% Hausdorff distance (HD95) as evaluation metrics. Higher Dice scores indicate better segmentation quality, whereas lower HD95 values represent more accurate boundary delineation. In the comparison results, the best and second-best performance values are highlighted in \textbf{bold} and \underline{underlined}.

\subsection{Comparison with State-of-the-Art Methods}

To comprehensively evaluate the performance of our MSM-Seg framework, we conduct extensive comparisons with state-of-the-art methods on multi-modal brain tumor segmentation across two challenging datasets from the BraTS 2024 challenge: BraTS-AGPT and BraTS-METS. All comparison baselines are reimplemented using the same training setting, as shown in Table~\ref{tab:compar}. \cz{The classical multi-modal segmentation methods \cite{wang2021transbts, yang20233d, shaker2024unetr++, isensee2024nnu} generally underperform compared to recent Mamba/RWKV-based variants \cite{xing2025segmamba, yu2025superlightnet}. Notably, our MSM-Seg framework operating in automatic mode (without manual prompts) significantly outperforms these baselines, achieving 75.01\% and 78.71\% average Dice scores on the BraTS-METS and BraTS-AGPT datasets, representing improvements of 1.09\% and 1.55\% over SuperLightNet with P-value $<$ 0.005. The HD95 results further demonstrate superior boundary precision, with MSM-Seg achieving 19.29mm and 7.02mm compared to SegMamba-V2's 20.40mm and SuperLightNet's 7.90mm, indicating better delineation of tumor boundaries across both metastases and glioma segmentation tasks.}

For prompt-based methods leveraging foundation model capabilities, we evaluate against various SAM-based approaches \cite{kirillov2023segment, chen2024ma, ravisam, zhu2024medical, chen2025sam2, yan2025samed}. \cz{On BraTS-METS, SAMed-2 \cite{yan2025samed} delivers competitive results with 77.47\% average Dice, while SAM-Med3D \cite{wang2025sam} achieves 82.48\% Dice on BraTS-AGPT. However, these methods rely on category-specific prompts per subregion. In contrast, our MSM-Seg framework with category-agnostic prompts substantially surpasses all SAM-based methods, achieving exceptional performance of 79.51\% and 83.84\% average Dice scores on the respective datasets, representing remarkable improvements of 2.04\% and 1.36\% over SAM-Med3D.} The HD95 metrics consistently validate superior boundary precision, with MSM-Seg achieving 13.75mm and 5.56mm compared to SAMed-2's 14.27mm and 6.12mm, demonstrating enhanced capability in precise tumor boundary delineation. These comprehensive comparisons validate the superior performance of our MSM-Seg framework on multi-modal brain tumor segmentation through modality-and-slice memory modeling with category-agnostic prompting.

\begin{table*}[!t]
    \centering
    \scriptsize
    \caption{\cz{Overhead analysis of modality memory capacity on BraTS-METS and BraTS-AGPT in the automatic mode.}}
    \setlength\tabcolsep{2.6pt}
    \resizebox{\textwidth}{!}{\begin{tabular}{c|cccc|cccc|cccc|cccc|ccc}
    \hline
    \multirow{3}{*}{Modality Memory} & \multicolumn{8}{c|}{BraTS-METS} & \multicolumn{8}{c|}{BraTS-AGPT} & \multirow{3}{*}{Latency} & \multirow{3}{*}{Params} & \multirow{3}{*}{GPU Mem.}\\
    \cline{2-17}
    & \multicolumn{4}{c|}{Dice (\%) $\uparrow$} & \multicolumn{4}{c|}{HD95 (mm) $\downarrow$} & \multicolumn{4}{c|}{Dice (\%) $\uparrow$} & \multicolumn{4}{c|}{HD95 (mm) $\downarrow$} & & &\\
    \cline{2-17}
    & ET & TC & WT & Avg & ET & TC & WT & Avg & ET & TC & WT & Avg & ET & TC & WT & Avg & & &\\
    \hline
    $k=0$ & 64.87 & 70.35 & 72.18 & 71.13 & 21.45 & 22.13 & 22.85 & 22.14 & 62.18 & 66.74 & 84.32 & 75.83 & 7.89 & 8.56 & 9.21 & 8.55 & {3.86s} & 52.8M & 11.2GB\\ 
    $k=1$ & 66.52 & 71.98 & 73.81 & 72.77 & 20.24 & 20.89 & 21.35 & 20.83 & 63.71 & 68.29 & 85.58 & 77.03 & 7.18 & 7.82 & 8.47 & 7.82 & 3.96s & 52.8M & 11.5GB\\
    $k=2$ & 67.63 & 73.21 & 75.14 & 74.07 & 19.47 & 20.07 & 20.25 & 19.88 & 64.89 & 69.67 & 86.47 & 78.07 & 6.68 & 7.28 & 7.89 & 7.30 & 4.04s & 52.8M & 11.7GB\\
    $k=3$ & 68.53 & 74.11 & 75.89 & 75.01 & 18.57 & 19.19 & 19.40 & 19.29 & 65.61 & 70.32 & 87.10 & 78.71 & 6.17 & 6.74 & 7.29 & 7.02 & 4.17s & 52.8M & 12.0GB\\
    \hline
    \end{tabular}}
    \label{tab:mc}
\end{table*}

\begin{figure*}[!t]
\centering
  \includegraphics[width=1\linewidth]{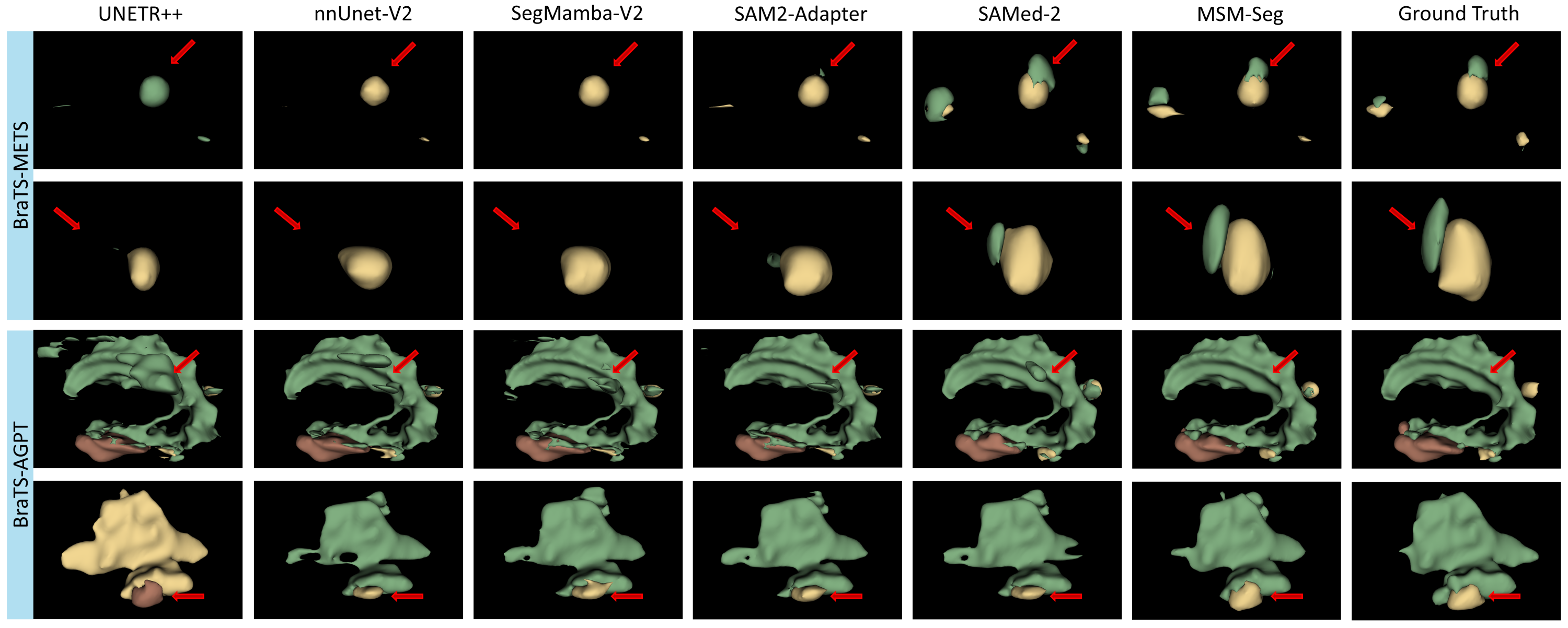}
  \caption{The 3D visualization of multi-modal brain tumor segmentation on the BraTS-METS and BraTS-AGPT datasets. Our MSM-Seg exhibits smoother surface boundaries, reduced fragmentation, and better preservation of complex 3D tumor morphology compared to state-of-the-art methods.}
  \label{fig:visual2}
\end{figure*}

\begin{figure}[!t]
    \centering
    \includegraphics[width=1\linewidth]{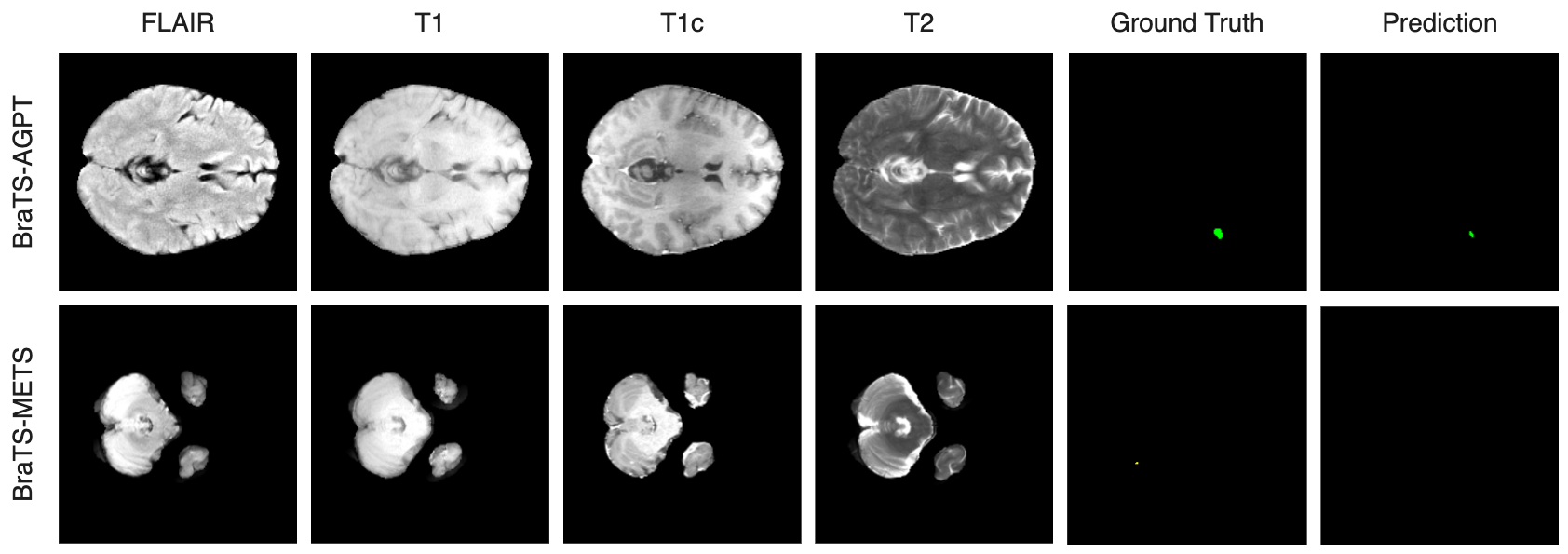}
    \caption{\cz{Representative failure cases of MSM-Seg, including weak-boundary under-segmentation and missed detection of tiny low-contrast lesions.}}
    \label{fig:failure}
\end{figure}

\subsection{Visualization Analysis}
To further demonstrate the effectiveness of our MSM-Seg framework, we provide qualitative results of both BraTS-METS and BraTS-AGPT datasets. As shown in Fig. \ref{fig:visual1}, we provide the visualization results of metastases and glioma tumors with three different subregions: enhancing tumor (yellow), surrounding non-enhancing FLAIR hyperintensity (green), and non-enhancing tumor core (red). Our MSM-Seg achieves exceptional performance in glioma and metastases tumor segmentation, precisely delineating distinct tumor subregions, including enhancing tumor, tumor core, and whole tumor areas. The category-agnostic prompting mechanism enables robust segmentation across varying tumor grades and morphological characteristics, while the modality-adaptive fusion effectively integrates complementary information from different MRI sequences. Furthermore, as shown in Fig. \ref{fig:visual2}, the 3D volumetric visualization reveals that MSM-Seg maintains exceptional spatial coherence across consecutive slices, producing smoother surface reconstructions with minimal fragmentation artifacts. These additional results further validate the superiority of our proposed MSM-Seg framework in multi-modal brain tumor segmentation, consistently outperforming existing methods in terms of boundary precision, structural completeness, and cross-modal consistency. 

\cz{It is worth noting that MSM-Seg still has limitations in challenging cases with insufficient visual evidence. As shown in Fig.~\ref{fig:failure}, the main failure patterns occur in extremely small or weakly contrasted tumor regions. In the BraTS-AGPT case, MSM-Seg captures the approximate tumor location, but the predicted region is smaller than the ground truth. In the BraTS-METS case, the lesion is extremely tiny and has limited contrast across modalities, and the model fails to produce a clear prediction. These cases indicate that MSM-Seg may still be limited for tiny lesions, weak boundaries, and ambiguous multi-modal appearances. Future work will investigate more effective small-lesion modeling and boundary refinement strategies to improve robustness in these challenging scenarios.}

\subsection{Ablation Study}
To investigate the effectiveness of MSMA $\mathcal{A}$, MCP-Encoder $\mathcal{E}$, MF-Decoder $\mathcal{D}$, and dual-memory segmentation paradigm $\mathcal{P}$, we conduct comprehensive ablation studies on multi-modal brain tumor segmentation across two challenging datasets, as illustrated in Table~\ref{tab:ab}. By removing the tailored modules from MSM-Seg, we construct an original SAM2 \cite{ravisam} framework with Hiera-S as the ablation baseline ($1^{st}$ row), achieving average Dice scores of 75.18\% and 78.55\% on BraTS-METS and BraTS-AGPT, respectively. By separately introducing the MSMA ($2^{nd}$ row), MCP-Encoder ($3^{rd}$ row), and MF-Decoder ($4^{th}$ row), the performance demonstrates consistent improvements across both datasets and all tumor regions. Specifically, the MSMA provides immediate improvements with average Dice gains of 0.65\% and 0.81\% on BraTS-METS and BraTS-AGPT, respectively, demonstrating the value of capturing complementary information between modalities and slices. The MCP-Encoder contributes additional improvements of 0.87\% and 1.07\% in average Dice, validating the effectiveness of category-agnostic multi-scale prompt encoding. The MF-Decoder further enhances performance with gains of 1.08\% and 1.33\%, confirming the benefits of modality-adaptive fusion for multi-modal tumor understanding. Notably, our dual-memory segmentation paradigm ($5^{th}$ row) achieves the most significant individual contribution, with substantial improvements averaging 1.73\% and 2.08\% in Dice scores on BraTS-METS and BraTS-AGPT, respectively. The full MSM-Seg attains optimal performance with average Dice scores of 79.51\% and 83.84\%, representing improvements of 4.33\% and 5.29\% over the baseline, respectively. These comprehensive ablation experiments with P-value $<$ 0.001 validate that the tailored MSMA, MCP-Encoder, MF-Decoder, and dual-memory segmentation paradigm collectively contribute to the superior performance of our MSM-Seg framework by effectively exploiting cross-modal and inter-slice relationships, providing efficient category-agnostic guidance.

\begin{table*}[!t]
    \centering
    \scriptsize
    \caption{\cz{Overhead analysis of slice memory capacity on BraTS-METS and BraTS-AGPT in the automatic mode without manual prompts.}}
    \setlength\tabcolsep{2.4pt}
    \scalebox{1.1}{\begin{tabular}{c|cccc|cccc|cccc|cccc|ccc}
    \hline
    \multirow{3}{*}{Slice Memory} & \multicolumn{8}{c|}{BraTS-METS} & \multicolumn{8}{c|}{BraTS-AGPT} & \multirow{3}{*}{Latency} & \multirow{3}{*}{Params} & \multirow{3}{*}{GPU Mem.}\\
    \cline{2-17}
    & \multicolumn{4}{c|}{Dice (\%) $\uparrow$} & \multicolumn{4}{c|}{HD95 (mm) $\downarrow$} & \multicolumn{4}{c|}{Dice (\%) $\uparrow$} & \multicolumn{4}{c|}{HD95 (mm) $\downarrow$} & & &\\
    \cline{2-17}
    & ET & TC & WT & Avg & ET & TC & WT & Avg & ET & TC & WT & Avg & ET & TC & WT & Avg & & &\\
    \hline
    $n=0$ & 62.84 & 68.23 & 70.45 & 69.17 & 22.73 & 23.52 & 24.18 & 23.48 & 60.29 & 64.58 & 82.57 & 73.81 & 8.94 & 9.67 & 10.35 & 9.65 & {3.49s} & 52.8M & 10.2GB\\ 
    $n=1$ & 64.58 & 70.12 & 72.07 & 70.59 & 21.42 & 22.15 & 22.74 & 22.10 & 61.76 & 66.29 & 83.94 & 75.33 & 8.13 & 8.82 & 9.47 & 8.81 & 3.58s & 52.8M & 10.4GB\\
    $n=2$ & 65.89 & 71.63 & 73.42 & 72.31 & 20.38 & 21.06 & 21.58 & 21.01 & 63.04 & 67.82 & 85.13 & 76.66 & 7.49 & 8.14 & 8.73 & 8.12 & 3.65s & 52.8M & 10.6GB\\
    $n=4$ & 67.05 & 72.74 & 74.52 & 73.44 & 19.57 & 20.21 & 20.67 & 20.15 & 64.19 & 69.01 & 86.02 & 77.57 & 6.95 & 7.56 & 8.14 & 7.55 & 3.82s & 52.8M & 11.1GB\\
    $n=8$ & 68.53 & 74.11 & 75.89 & 75.01 & 18.57 & 19.19 & 19.40 & 19.29 & 65.61 & 70.32 & 87.10 & 78.71 & 6.17 & 6.74 & 7.29 & 7.02 & 4.17s & 52.8M & 12.0GB\\
    $n=16$ & 69.17 & 74.83 & 76.74 & 75.91 & 17.94 & 18.53 & 18.81 & 18.62 & 66.28 & 71.17 & 88.05 & 79.50 & 5.81 & 6.35 & 6.92 & 6.63 & 5.03s & 52.8M & 13.5GB\\
    \hline
    \end{tabular}}
    \label{tab:sc}
\end{table*}

\begin{table*}[!t]
    \centering
    \small
    \caption{\cz{Analysis of modality-adaptive fusion prediction on BraTS-METS and BraTS-AGPT in the category-agnostic prompt mode using whole tumor bounding box prompts.}}
    \setlength\tabcolsep{3.7pt}
    \scalebox{1}{\begin{tabular}{cccc|cccc|cccc|cccc|cccc}
    \hline
    \multirow{3}{*}{T2} & \multirow{3}{*}{T1c} & \multirow{3}{*}{T1} & \multirow{3}{*}{FLAIR} & \multicolumn{8}{c|}{BraTS-METS} & \multicolumn{8}{c}{BraTS-AGPT}\\
    \cline{5-20}
    & & & & \multicolumn{4}{c|}{Dice (\%) $\uparrow$} & \multicolumn{4}{c|}{HD95 (mm) $\downarrow$} & \multicolumn{4}{c|}{Dice (\%) $\uparrow$} & \multicolumn{4}{c}{HD95 (mm) $\downarrow$}\\
    \cline{5-20}
    & & & & ET & TC & WT & Avg & ET & TC & WT & Avg & ET & TC & WT & Avg & ET & TC & WT & Avg\\
    \hline
    \usym{1F5F8} &  &  &  & 66.73 & 73.82 & 76.94 & 74.83 & 17.29 & 16.47 & 15.68 & 16.48 & 64.58 & 68.95 & 87.42 & 78.32 & 7.05 & 6.31 & 5.56 & 6.31\\
    \usym{1F5F8} & \usym{1F5F8} &  &  & 68.47 & 76.24 & 78.58 & 76.43 & 16.12 & 15.38 & 14.67 & 15.39 & 65.82 & 70.76 & 89.36 & 80.31 & 6.51 & 5.82 & 5.31 & 5.88\\
    \usym{1F5F8} & \usym{1F5F8} & \usym{1F5F8} &  & 69.87 & 77.65 & 79.82 & 77.78 & 15.41 & 14.73 & 14.08 & 14.74 & 66.42 & 71.94 & 90.73 & 81.70 & 6.13 & 5.49 & 5.22 & 5.58\\
    \usym{1F5F8} & \usym{1F5F8} & \usym{1F5F8} & \usym{1F5F8} & 71.82 & 79.45 & 81.27 & 79.51 & 14.67 & 13.84 & 12.73 & 13.75 & 67.29 & 73.76 & 92.43 & 83.84 & 5.94 & 5.21 & 5.15 & 5.56\\
    \hline
    \end{tabular}}
    \label{tab:dec}
\end{table*}

\begin{table*}[!t]
    \centering
    \scriptsize
    \caption{\cz{Tumor-size stratified comparison with state-of-the-art prompt-based segmentation methods on BraTS-METS and BraTS-AGPT in the category-agnostic prompt mode using whole tumor bounding box prompts.}}
    \setlength\tabcolsep{3.0pt}
    \renewcommand{\arraystretch}{0.98}
    \resizebox{\textwidth}{!}{\begin{tabular}{l|cc|cc|cc|cc|cc|cc|cc|cc}
    \hline
    \multirow{3}{*}{Method} 
    & \multicolumn{8}{c|}{BraTS-METS} 
    & \multicolumn{8}{c}{BraTS-AGPT} \\
    \cline{2-17}
    & \multicolumn{2}{c|}{Small} 
    & \multicolumn{2}{c|}{Medium} 
    & \multicolumn{2}{c|}{Large} 
    & \multicolumn{2}{c|}{Overall}
    & \multicolumn{2}{c|}{Small} 
    & \multicolumn{2}{c|}{Medium} 
    & \multicolumn{2}{c|}{Large} 
    & \multicolumn{2}{c}{Overall} \\
    \cline{2-17}
    & Dice & HD95 & Dice & HD95 & Dice & HD95 & Dice & HD95
    & Dice & HD95 & Dice & HD95 & Dice & HD95 & Dice & HD95 \\
    \hline
    SAM2-Adapter \cite{chen2025sam2}
    & 71.24 & 18.93 & 76.84 & 14.71 & 82.32 & 10.61 & 76.80 & 14.75
    & 75.38 & 8.24 & 80.92 & 6.19 & 86.58 & 4.08 & 80.96 & 6.17 \\
    SAMed-2 \cite{yan2025samed}
    & 71.87 & 18.42 & 77.51 & 14.26 & 83.04 & 10.13 & 77.47 & 14.27
    & 75.91 & 8.07 & 81.45 & 6.09 & 87.02 & 3.94 & 81.44 & 6.12 \\
    MSM-Seg 
    & 74.65 & 17.35 
    & 79.77 & 13.56 
    & 84.11 & 9.73 
    & 79.51 & 13.75
    & 78.54 & 7.15 
    & 83.90 & 5.53 
    & 89.08 & 3.55 
    & 83.84 & 5.56 \\
    \hline
    \end{tabular}}
    \label{tab:size}
\end{table*}

\begin{table*}[!t]
    \centering
    \scriptsize
    \caption{\cz{Sensitivity analysis of prompt perturbations on BraTS-METS and BraTS-AGPT in the category-agnostic prompt mode.}}
    \setlength\tabcolsep{2.8pt}
    \resizebox{\textwidth}{!}{\begin{tabular}{c|cccc|cccc|cccc|cccc}
    \hline
    \multirow{3}{*}{Prompt Perturbation} & \multicolumn{8}{c|}{BraTS-METS} & \multicolumn{8}{c}{BraTS-AGPT}\\
    \cline{2-17}
    & \multicolumn{4}{c|}{Dice (\%) $\uparrow$} & \multicolumn{4}{c|}{HD95 (mm) $\downarrow$} & \multicolumn{4}{c|}{Dice (\%) $\uparrow$} & \multicolumn{4}{c}{HD95 (mm) $\downarrow$}\\
    \cline{2-17}
    & ET & TC & WT & Avg & ET & TC & WT & Avg & ET & TC & WT & Avg & ET & TC & WT & Avg\\
    \hline
    $R=0$ px  & 71.82 & 79.45 & 81.27 & 79.51 & 14.67 & 13.84 & 12.73 & 13.75 & 67.29 & 73.76 & 92.43 & 83.84 & 5.94 & 5.21 & 5.15 & 5.56\\
    $R=5$ px  & 71.38 & 79.08 & 80.94 & 79.13 & 14.98 & 14.14 & 13.04 & 14.05 & 66.95 & 73.43 & 92.16 & 83.51 & 6.09 & 5.34 & 5.28 & 5.67\\
    $R=10$ px & 70.74 & 78.47 & 80.28 & 78.71 & 15.54 & 14.81 & 13.58 & 14.64 & 66.32 & 72.87 & 91.67 & 83.07 & 6.31 & 5.55 & 5.47 & 5.78\\
    $R=20$ px & 69.17 & 77.12 & 79.21 & 77.17 & 16.89 & 16.02 & 14.81 & 15.91 & 65.47 & 71.98 & 90.83 & 82.09 & 6.68 & 5.89 & 5.83 & 6.13\\
    \hline
    \end{tabular}}
    \label{tab:box_sensitivity}
\end{table*}

\subsection{Modality Sequence Analysis}
To investigate the impact of modality processing order on segmentation performance, we evaluate different arrangements of the four MRI modalities (FLAIR, T1, T1c, T2) across both datasets, as shown in Table~\ref{tab:modality_sequence}. On BraTS-METS, the default sequence (FLAIR→T1→T1c→T2) achieves 79.51\% average Dice, while alternative sequences show minimal variations within 0.13\% for Dice and 0.12mm for HD95. Similarly, on BraTS-AGPT, the default sequence achieves 83.84\% average Dice with alternative arrangements demonstrating fluctuations within 0.30\%. Statistical t-test analysis reveals no significant differences (P-values $>$ 0.05) across different modality orders on both datasets. This robustness stems from our modality-and-slice memory attention mechanism, which captures cross-modal complementary relationships regardless of input order, and our dual-memory paradigm that synergistically integrates multi-modal information. Unlike traditional methods that exhibit sensitivity to modality ordering due to sequential processing or fixed fusion strategies, our sequence-agnostic property offers practical advantages for clinical deployment, as different medical centers may acquire MRI sequences in varying orders based on scanner protocols or patient conditions.

\subsection{Modality-and-Slice Memory Designs}
To investigate the impact of memory mechanisms on segmentation performance, we analyze modality memory capacity ($k$) and slice memory capacity ($n$) in automatic mode across BraTS-METS and BraTS-AGPT datasets. \cz{To provide a detailed performance-efficiency analysis, we additionally report parameter count and GPU memory usage for each variant.} As shown in Table~\ref{tab:mc}, without modality memory ($k=0$), the framework processes each modality independently, achieving baseline average Dice scores of 71.13\% and 75.83\% with HD95 of 22.14mm and 8.55mm. \cz{The performance improves progressively with increasing $k$, reaching optimal results at $k=3$ with 75.01\% and 78.71\% average Dice (improvements of 3.88\% and 2.88\%) and HD95 reduced to 19.29mm and 7.02mm, demonstrating the effectiveness of capturing cross-modal complementary information.} \cz{Across $k=0$ to $k=3$, MSM-Seg maintains the same parameter count of 52.8M with only a modest GPU memory increase from 11.2GB to 12.0GB.} The inference latency increases modestly from 3.86s to 4.17s (8.0\% overhead), indicating reasonable computational cost for these significant gains.

Table~\ref{tab:sc} reveals that slice memory capacity has a more substantial impact on segmentation performance. Without slice memory ($n=0$), the model achieves only 69.17\% and 73.81\% average Dice with HD95 of 23.48mm and 9.65mm. Performance scales consistently with $n$, and $n=16$ achieves optimal scores of 75.91\% and 79.50\% (improvements of 6.74\% and 5.69\% over baseline). \cz{The larger gains from slice memory compared to modality memory (6.74\% vs. 3.88\% on BraTS-METS) indicate that spatial context propagation plays a more dominant role in our dual-memory framework.} \cz{However, from $n=8$ to $n=16$, marginal gains (0.90\% and 0.79\%) come at significantly increased latency (4.17s to 5.03s, 20.6\% overhead) and GPU memory usage (12.0GB to 13.5GB).} \cz{The parameter count remains unchanged at 52.8M across different slice-memory capacities, since varying $n$ only changes the number of stored memory features rather than learnable parameters.} Considering the performance-efficiency trade-off, we adopt $k=3$ and $n=8$ as default settings for MSM-Seg. \cz{These analyses validate that our dual-memory design effectively captures both cross-modal dependencies and spatial continuity of tumor structures with acceptable computational overhead.}

\subsection{Effectiveness of Modality-Adaptive Fusion}
To evaluate our modality-adaptive fusion strategy, we progressively incorporate different MRI modalities in the MF-Decoder, as shown in Table~\ref{tab:dec}. Using only T2, the framework achieves baseline average Dice scores of 74.83\% and 78.32\% on BraTS-METS and BraTS-AGPT, respectively. Sequentially adding T1c and T1 yields cumulative improvements, with T1c contributing 1.60\%/1.99\% and T1 adding 1.35\%/1.39\% in average Dice. The complete four-modality fusion, including FLAIR for delineating peritumoral edema, achieves optimal performance with 79.51\% and 83.84\% average Dice scores, representing substantial gains of 4.68\% and 5.52\% over single-modality segmentation, along with HD95 reductions of 2.73mm and 0.75mm. Notably, the non-linear performance gains across modality combinations indicate that our adaptive fusion mechanism captures synergistic cross-modal relationships rather than simply averaging predictions. This design also offers practical flexibility for clinical scenarios where certain MRI sequences may be unavailable.

\subsection{Tumor-Size Stratified Analysis}

\cz{To further evaluate the robustness of MSM-Seg across different lesion scales, we conduct a tumor-size stratified analysis on BraTS-METS and BraTS-AGPT. For each test sample, we calculate the tumor size according to the ground truth whole tumor region and divide the cases into small ($<$5 mm), medium (5--10 mm), and large ($>$10 mm) groups. As shown in Table~\ref{tab:size}, MSM-Seg consistently outperforms SAM2-Adapter and SAMed-2 across all tumor-size groups on both datasets. Compared with the strongest baseline SAMed-2, MSM-Seg improves the average Dice by 2.78\%, 2.26\%, and 1.07\% for small, medium, and large tumors on BraTS-METS, and by 2.63\%, 2.45\%, and 2.06\% on BraTS-AGPT, respectively. Meanwhile, MSM-Seg reduces the average HD95 by 1.07 mm, 0.70 mm, and 0.40 mm on BraTS-METS, and by 0.92 mm, 0.56 mm, and 0.39 mm on BraTS-AGPT. These results demonstrate that MSM-Seg maintains consistent advantages across different tumor sizes, especially for challenging small tumors with limited spatial extent and weak appearance cues.}

\subsection{Bounding Box Prompt Sensitivity Analysis}

\cz{To evaluate the robustness to inaccurate category-agnostic prompts, we conduct a bounding box prompt sensitivity analysis of MSM-Seg. Following the MedSAM prompt perturbation protocol, we first generate the tight bounding box from the ground truth whole tumor mask as the original prompt. To simulate inaccurate user-provided boxes, each box boundary is independently shifted by a random value sampled from $[-R, R]$, where $R$ is set to 5, 10, or 20 pixels. As shown in Table~\ref{tab:box_sensitivity}, MSM-Seg degrades gradually as the perturbation range increases. On BraTS-METS, the average Dice decreases from 79.51\% to 79.13\%, 78.71\%, and 77.17\% under $R=5$, $R=10$, and $R=20$ pixels, respectively, while the average HD95 increases from 13.75 mm to 14.05 mm, 14.64 mm, and 15.91 mm. On BraTS-AGPT, the average Dice decreases from 83.84\% to 83.51\%, 83.07\%, and 82.09\%, and the average HD95 increases from 5.56 mm to 5.67 mm, 5.78 mm, and 6.13 mm. These results indicate that MSM-Seg remains reasonably robust to moderate inaccuracies in category-agnostic whole tumor bounding box prompts.}

\section{Conclusion}
In this work, we propose the MSM-Seg, a synergistic cross-modal framework tailored for brain tumor segmentation. \cz{The proposed MSM-Seg framework integrates a MSMA to exploit cross-modal and inter-slice relationships, a MCP-Encoder that provides whole tumor region guidance through an efficient category-agnostic design, and a MF-Decoder that leverages complementary decoding information across modalities to improve segmentation accuracy.} Extensive experiments on multi-modal 3D brain glioma and metastases datasets demonstrate that our MSM-Seg framework outperforms state-of-the-art methods in 3D multi-modal brain tumor segmentation.

\balance
\bibliographystyle{IEEEtran}
\bibliography{refs}

\vfill

\end{document}